# Double-Coupling Learning for Multi-Task Data Stream Classification

Yingzhong Shi, College of Internet of Things, Wuxi Institute of Technology, Wuxi, China; School of Digital Media, Jiangnan University, Wuxi, China
Zhaohong Deng, School of Digital Media, Jiangnan University, Wuxi, China
HaoranChen, Mathematics Department, University of Michigan, Ann Arbor, USA
Kup-Sze Choi, Center of Smart Heath, Hong Kong Polytechnic University, Hong Kong
Shitong Wang, School of Digital Media, Jiangnan University, Wuxi, China

*Abstract*—Data stream classification methods demonstrate promising performance on a single data stream by exploring the cohesion in the data stream. However, multiple data streams that involve several correlated data streams are common in many practical scenarios, which can be viewed as multi-task data streams. Instead of handling them separately, it is beneficial to consider the correlations among the multi-task data streams for data stream modeling tasks. In this regard, a novel classification method called double-coupling support vector machines (DC-SVM), is proposed for classifying them simultaneously. DC-SVM considers the external correlations between multiple data streams, while handling the internal relationship within the individual data stream. Experimental results on artificial and real-world multi-task data streams demonstrate that the proposed method outperforms traditional data stream classification methods.

Multi-task Data Streams, Classification, Internal and External Coupling, Concept Drift

## I. INTRODUCTION

For the modeling of multiple tasks that are interrelated, it is advantageous to learn the model for all the tasks simultaneously instead of handling each task independently. Many studies have demonstrated the benefits of multi-task learning over individual task learning if the tasks are related [1-3]. The principal objective of multi-task learning is to improve the generalization performance by leveraging domain specific information that is contained in the related tasks [1]. The training signals from the extra tasks serve as an inductive bias that is helpful for learning the multiple complex tasks together [1,2]. Empirical and theoretical studies on multi-task learning have been conducted in three areas: multi-task classification [1-5], multi-task clustering [6-8] and multi-task regression [9,10]. Although having demonstrated the significance of multi-task learning in various real-world applications, the current multi-task learning methods are limited to stationary patterns and cannot keep up with some new requirements, particularly on dynamic data and data streams. In this study, we focus on multi-task data stream classification.

A data stream is a continuous and changing sequence of data that arrives at a system where the data are stored or processed [11-14]. Differs from traditional data, data stream is dynamic, evolving, high-speed, ordered, and infinite-length. Typical examples of data streams include weather forecasting data, sensor data, and so on. Three main techniques of data stream mining are classification, clustering and frequent pattern mining. Classification is an important data stream mining technique, which aims at predicting an independent variable (class label) according to values of an instance. In real-world applications, data are not always sufficient for constructing a reliable classification model for a data stream at a specified time. Therefore, the effective use of the related information in adjacent times is crucial. Most previous approaches use local classifiers, each of which fits a time window of a specified duration [15-20]. These methods differ in the selection of the window length, the setting of the weights of the selected samples, or the assembly of the set of classifiers. The time-adaptive support vector machine (TA-SVM) [19] is a representative work, which can simultaneously deal with a sequence of subclassifiers and compromise between local optimality (individual optimality) and global optimality (adjacent similarity). Furthermore, the improved time-adaptive support vector machine (ITA-SVM) was proposed in [20] for realizing asymptotic linear time complexity for large nonstationary datasets.

Traditional data stream classification methods mainly focus on individual data streams. However, many practical scenarios involve several related data streams. For example, weather forecasting data streams of neighboring cities may be related, website customer click streams in several portals may be associated, the measured values of gas sensor array data for different gases may drift synchronously when the performance of the gas sensor arrays deteriorates [21], water quality measurements in neighboring areas have similarities and may drift synchronously, and the concentrations of pollutants that have similar chemical properties may have the same variational trend [48]. The above situations have a common characteristic that multiple data streams for different tasks have similar correlation and different tasks may benefit from each other.

Thus, useful information may be lost if the individual data streams are only considered in the above scenarios. It is necessary to develop modeling methods that can exploit several data streams simultaneously.

In this study, the multiple data stream modeling task for classification is considered, which is a multi-task data stream modeling problem. Multi-task data stream classification faces two main challenges: (1) from the viewpoint of data stream modeling, more attention should be paid to the relationships between data streams; (2) from the viewpoint of multi-task learning, data evolution in each task must be handled, which is not considered in traditional multi-task learning. To the best of our knowledge, few works have focused on these two challenges simultaneously. In this paper, a novel multiple data streams classification method, namely, double-coupling support vector machines (DC-SVM), is proposed for the above challenges.

The proposed DC-SVM has the following characteristics: (1) Internal coupling is used within the individual data stream such that the classifiers in the adjacent time windows are constrained to be similar. (2) External coupling between the related data streams is introduced such that at a specified time, the classifiers of several related data streams are constrained to be similar. (3) DC-SVM can process several data streams simultaneously and outperforms the traditional methods, such as TA-SVM and ITA-SVM, which process data streams independently. (4) DC-SVM inherits the formulation of SVM in an alternative kernel space, which make many efficient optimization techniques for SVM available to DC-SVM directly.

The remainder of this paper is organized as follows: In Section II, the related work is briefly reviewed. In Section III, the details of DC-SVM is presented. Extensive experimental studies are given in Section IV. Finally, the conclusions of are presented in Section V.

## II. RELATED WORK

In data stream classification, concept drift [11,12] is the main issue that affects the classification performance. Concept drift can be classified into two categories: gradual concept drift [22-27] and sudden drift [28-32]. Besides data stream classification, research on data stream clustering [33-35], data stream feature selection [36-37] and other scenes [38-41] have also received attention.

This paper considers data stream classification, with a focus on gradual concept drift in which the patterns are assumed to change slowly such that rapid reaction is not necessary. Most previous approaches that tackle concept drift rely on the use of "local" classifiers, each fitting or adapting to a sliding time window of a specified length [14,16-18,28]. Some researchers prefer to use the equivalent idea of uniform or stationary "batches" [18,42-43]. Nevertheless, the sliding window methods all face the following dilemma: If the time window is too short, the captured information will be insufficient. In contrast, if the time window is too long, the concept drift could become significant, thereby reducing the reliability of the classification model. A potential solution is to utilize an adaptive window size and some methods have been proposed [14, 16, 17, 19, 20, 44, 45].

TA-SVM and ITA-SVM are two data stream classification methods, which are very related with our work in this study. Here, they are introduced briefly. Both TA-SVM and ITA-SVM consider the local optimization of each subclassifier, with the assumption that the changes between subsequent subclassifiers are relatively stable over time. In TA-SVM, the differences between the subclassifiers are constrained as follows:

$$\min \sum_{t=1}^{T} \|f_t\|^2 + \gamma \sum_{t=1}^{T-1} \|f_{t+1} - f_t\|^2 \tag{1}$$

where $\|f_t\|^2$ is the local regularization term of the $t$th subclassifier $f_t$; $f_{t+1} - f_t$ is the discriminating between two adjacent subclassifiers; $\sum_{t=1}^{T-1} \|f_{t+1} - f_t\|^2$ is the global optimization term; and $\gamma$ is a parameter for balancing the local and global optimizations. ITA-SVM, as a improved version of TA-SVM, aims at solving for a common term and a set of incremental sequences by balancing the local optimization and the global optimization. The primal optimization of ITA-SVM is designed as follows:

$$\min \|f_0\|^2 + \gamma \sum_{t=1}^{T-1} \|\Delta f_t\|^2 \tag{2}$$

where $f_0$ is the shared term for all subclassifiers; $\Delta f_t = f_{t+1} - f_t$ is the increment of the two adjacent subclassifiers; $\sum_{t=1}^{T-1} \|\Delta f_t\|^2$ is the global optimization term; and $\gamma$ is the compromise factor for balancing the influences of the two terms. The time complexity of ITA-SVM is asymptotically linear, which is highly suitable for large-scale nonstationary datasets.

As described above, the existing data stream classification methods are mainly developed for a single data stream. For multiple related data streams, these methods only handle each data stream separately. They are not able to consider the potentially useful information among the related data streams, which can be beneficial for the classification of all data streams. Therefore, it is significant to investigate the specified classification methods for multi-task data streams.

## III. DOUBLE-COUPLING SVM FOR MULTI-TASK DATA STREAMS

In this section, we will present a method for multi-task data streams classification. First, the term "*multi-task data streams*" is defined as follows.

**Definition**: If two or more data streams are related, similar at various time windows, and the degree of similarity between them remains stable, then these data streams are called multi-task data streams.

For handling multi-task data streams simultaneously effectively, a method, called double-coupling (internal and external coupling) support vector machine (DC-SVM) is proposed. DC-SVM is developed based on the following assumptions: 1) local optimization, 2) internal-coupling and 3) external-coupling. The first one assumes that the current model should adapt to the current sampling data. The second one assumes that the model should change smoothly and the models associated with the adjacent times should have high correlation and similarity. The third one assumes that at each moment, the models for the multi-task data streams are similar and related. Fig. 1 shows the mechanism of DC-SVM. In the figure, the circles and ellipses represent sub-datasets at different time windows of data streams associated with Task1 and Task2, respectively. The two colors in a circle represent two classes in the dataset. $f_t$ and $g_t$ denote the classification models at time $t$ for data streams of Task1 and Task2, respectively.

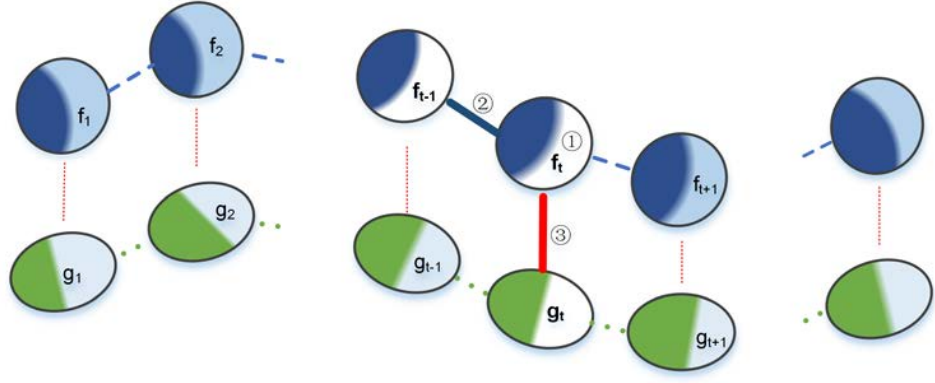

Fig. 1 Strategy of DC-SVM

As illustrated in Fig. 1, 1) $f_t$ adapts to the current sub-dataset in the corresponding circle, 2) $f_t$ is similar to its precursor $f_{t-1}$, and 3) $f_t$ is similar to its neighbor $g_t$. The main strategy of DC-SVM can be expressed by the following optimization problem:

$$\min \sum_{t=1}^{T}\|f_t\|^2 + \gamma_1 \sum_{t=1}^{T-1}\|f_{t+1} - f_t\|^2 + \sum_{t=1}^{T}\|g_t\|^2 + \gamma_2 \sum_{t=1}^{T-1}\|g_{t+1} - g_t\|^2 + \lambda \sum_{t=1}^{T}\|f_t - g_t\|^2 + C \sum L(f, g, \mathbf{x}, y) \tag{3}$$

where $\|f_t\|^2$ and $\|g_t\|^2$ are the local regularization terms of the $t$th subclassifiers $f_t$ and $g_t$, respectively; $\min \sum_{t=1}^{T-1}\|f_{t+1} - f_t\|^2$ and $\min \sum_{t=1}^{T-1}\|g_{t+1} - g_t\|^2$ are used to constrain the similarity within data streams Task1 and Task2, respectively; and $\gamma_1, \gamma_2$ are the compromise factors. The term $\min \sum_{t=1}^{T}\|f_t - g_t\|^2$ is to measure the similarity between data streams of Task1 and Task2 at time window $t$ and $\lambda$ is used to control the influence of this term; $\sum L(f, g, \mathbf{x}, y)$ is the term of loss function.

Let the dataset of a multi-task data stream be $\{(\mathbf{x}_i, \mathbf{y}_i) | i = 1, 2, \cdots, n\}$. The dataset contains $k$ individual data streams. Each data stream has $m$ time windows and each sub-dataset has $l$ samples, $n = k \times m \times l$. For each sub-dataset corresponds a sub-classifier, there are total $k \times m$ sub-classifiers. Denote $\mathbf{P} = [P_{\mu i}]_{(km) \times n}$ as a $(km) \times n$ matrix which is used to identify whether the $i$th sample belongs to the

$\mu$th sub-datasets, $\mu = 1, 2, \cdots, k, k+1, \cdots, km$, the sub-dataset is ordered by time firstly. $P_{\mu i} = 1$ if and only if $i \in p_\mu$; otherwise $P_{\mu i} = 0$. In this paper, we focus on multiple data streams that contain two related individual data streams, namely, $k = 2$, so $\mathbf{P}$ is a $2m \times n$ matrix. $\mathbf{R}^\gamma$ represents the internal-relationship of a single data stream, $\mathbf{R}^\gamma = [R^\gamma_{\mu\nu}]_{2m \times 2m}$, $R^\gamma_{\mu\nu} = 1$ if and only if $|\mu - \nu| = k$ ($k = 2$)(i.e., the $\mu$th classifier and $\nu$th classifier are adjacent in a single data stream); otherwise $R^\gamma_{\mu\nu} = 0$. $\mathbf{R}^\lambda = [R^\lambda_{\mu\nu}]_{2m \times 2m}$ represents the external-relationship between two data streams. The classifiers at the same time window can be viewed as neighbors. When $\mu, \nu$ indicate the same time, then $R^\lambda_{\mu\nu} = 1$; otherwise, $R^\lambda_{\mu\nu} = 0$. For example, $R^\lambda_{12} = R^\lambda_{21} = 1$, $R^\lambda_{23} = R^\lambda_{32} = 0$, $R^\lambda_{34} = R^\lambda_{43} = 1$. If $(\mathbf{w}_\mu, b_\mu)$ denote the $\mu$th classifier, the objective function of DC-SVM can be re-expressed in the following form:

$$\min_{w_\mu, b_\mu} \frac{1}{2m} \sum_{\mu=1}^{2m} \left( \|\mathbf{w}_\mu\|^2 + b_\mu^2 \right) + \frac{\gamma}{2m} \sum_{\mu=1}^{2m-2} \left( \|\mathbf{w}_{\mu+2} - \mathbf{w}_\mu\|^2 + (b_{\mu+2} - b_\mu)^2 \right)$$

$$+ \frac{\lambda}{2m} \sum_{\mu=1}^{m} \left( \|\mathbf{w}_{2\mu} - \mathbf{w}_{2\mu-1}\|^2 + (b_{2\mu} - b_{2\mu-1})^2 \right) - \rho + \frac{C}{2} \sum_{i=1}^{n} \xi_i^2$$

$$\text{s.t.} \quad y_i \left( \mathbf{w}^T_{\mu(i)} \varphi(\mathbf{x}_i) + b_{\mu(i)} \right) \geq \rho - \xi_i \tag{4}$$

Here, the flexible margin term $\rho$ is adopted as [20,46]. Based on dual optimization theory, the dual problem in (4) can be expressed as

$$\max_{\boldsymbol{\alpha}} -\frac{1}{2} \boldsymbol{\alpha}^T ((\mathbf{P}^T \mathbf{M}^{-1} \mathbf{P}) \otimes (\mathbf{K} + \mathbf{1}^T \mathbf{1}) \otimes \mathbf{Y} + \mathbf{I}/C) \boldsymbol{\alpha}$$

$$\text{s.t.} \quad \boldsymbol{\alpha} \geq 0, \quad \boldsymbol{\alpha}^T \mathbf{1} = 1. \tag{5}$$

Here, $\boldsymbol{\alpha} = (\alpha_1, \alpha_2, \cdots, \alpha_n)^T$, $\alpha_i$ is Lagrange Multiplier; $\mathbf{K} = [K_{ij}]_{n \times n}$, $K_{ij} = \varphi^T(\mathbf{x}_i) \varphi(\mathbf{x}_j)$, $\varphi$ is the kernel function; $\mathbf{Y} = [Y_{ij}]_{n \times n}$, $Y_{ij} = y_i \cdot y_j$; $\mathbf{I}$ is the identity matrix; $\mathbf{1} = (1_1, 1_2, \cdots, 1_n)^T$; $\mathbf{M} = [M_{\mu\nu}]_{2m \times 2m}$, $M_{\mu\nu} = (1 + \gamma \sum_k R^\gamma_{\mu k} + \lambda \sum_k R^\lambda_{\mu k})/m$ if $\mu = \nu$, otherwise $M_{\mu\nu} = (-\gamma R^\gamma_{\mu k} - \lambda R^\lambda_{\mu k})/m$. The notation $\otimes$ denotes the entrywise matrix product that if $\mathbf{A} = \mathbf{B} \otimes \mathbf{C}$ then $A_{ij} = B_{ij} \cdot C_{ij}$. According to (5), the optimization of DC-SVM is a QP problem, which can be solved using many available QP solution techniques. In the situation when $k > 2$, we can keep the Eq.(5) while adjust the matrix $\mathbf{P}, \mathbf{R}^\gamma, \mathbf{R}^\lambda$.

## IV. EXPERIMENTAL STUDY AND DISCUSSION

The performance of DC-SVM is experimentally investigated in this section. Since there are no previous studies on multi-task data streams, we adopted two single data stream classification methods, i.e., TA-SVM and ITA-SVM for comparison. When TA-SVM and ITA-SVM were used, the relationship among the multiple data streams was disregarded and each data stream was handled independently. In some scenarios, we combined several data streams into a new single data stream and solved it via ITA-SVM, which is denoted by ITA-SVM (Merge). All the experiments were conducted on a computer with a 3.30 GHz processor and 8 GB memory. All algorithms were realized using the MATLAB R2010a.

Table 1

Data streams that were used in the experiments

| Dataset | Description |
| --- | --- |
| DS1- Task1 | Sliding Gaussian data stream [19] |
| DS1- Task2 | Data stream DS1-Task1 with deviation; see (6) |
| DS2- Task1 | Rotating hyperplane [19] |
| DS2- Task2 | Rotating slightly from DS2-Task1 |
| DS3 | DS2 with half samples |
| DS4 | DS1 with 10% noise |
| DS5 | DS2 with 10% noise |
| GSADD | Gas sensor array drift dataset, UCI Dataset [21] |
| Water Quality | Water quality monitoring data stream [47] |
| Air Quality | UCI Dataset Air Quality [48] |

### A. Experimental Settings

Grid search was used for parameter optimization and the grids were set as follows: regularization parameter $C \in \{10^{-1}, 10^0, \cdots, 10^5\}$, Gaussian kernel $\sigma \in \{2^{-2}, 2^{-1}, 2^0, \cdots, 2^5\}$, internal-coupling parameter $\gamma \in \{2^0, 2^1, \cdots, 2^{20}\}$, external-coupling parameter $\lambda \in \{2^0, 2^1, \cdots, 2^{20}\}$.

The datasets that were used in the experiments are listed in Table 1. They are described as follows: *1)*

The artificial sliding Gaussian dataset that was used in [19] is denoted as DS1-Task1. Dataset DS1-Task2 was generated based on DS1-Task1 according to (6).

$$x_t = \left\{\frac{2t\pi}{n} - \pi + 0.2y_t + \varepsilon_1, (1-r)*\sin(\frac{2t\pi}{n} - \pi + 0.2y_t) + \varepsilon_2\right\} \tag{6}$$

Here, $t = 1, \cdots, n(n = 500)$; $\varepsilon_{1,2}$ are random perturbations that were sampled from a normal distribution with zero mean and a standard deviation of $\sigma = 0.1$; and $y_t$ is balanced random sequences of $\pm 1$. The parameter $r$ reflects the deviation of DS1-Task2 from DS1-Task1, where $r \in \{0.05, 0.1, 0.2, 0.3\}$. When $r = 0$, that is DS1-Task1. *2)* The rotating hyperplane dataset that was used in [19] is denoted as DS2-Task1, which is a commonly used dataset for analyzing data streams. Datasets DS2-Task2 shared the same pattern with DS2-Task1. The only difference between them was that there was an angle $r$ ($r \in \{2^0, 4^0, 6^0, 8^0, 10^0\}$) between their normal vectors. Thus, the similar data streams DS2-Task1 and DS2-Task2 can be used to compose the correlative data streams DS2. *3)* Data streams DS3 have the same distribution as DS2, but with half the sampling speed, namely, 250 samples are obtained from each of DS3-Task1 and DS3-Task2. DS3 is used to investigate the performance of DC-SVM in the state of data scarcity. *4)* Multi-task data streams DS4 and DS5 were generated by randomly switching the labels in DS1 and DS2, respectively, with a probability of 10%. They were used to evaluate the performance of DC-SVM in a noisy environment. *5)* Several real-world data streams, namely, gas sensor array drift dataset [21] and water quality [47] and air quality[48] datasets, were used to evaluate the performance of DC-SVM in practical applications. The details of these datasets are presented in subsections D, E and F, respectively.

*B. Analyses on Coupling Parameter $\lambda$ and $\gamma$*

Experiments were conducted to analyze the external-coupling parameter $\lambda$ and internal -coupling parameter $\gamma$ in the proposed method. The estimation of appropriate values for coupling parameters is crucial. Now, we consider the correlated data streams DS1-Task1 and DS1-Task2 with the different extent deviation between them to analyze how to set appropriate values for coupling parameters $\lambda$ and $\gamma$. Fix

$C=10$ and $\sigma=1$ (when the Gaussian kernel was adopted). The optimal values of $\lambda$ and $\gamma$ in terms of the performance of DC-SVM over 10 runs are recorded. Increased the deviation between them gradually with $r=0.1, 0.2, 0.3$ and record the best parameters over 10 runs. Fig. 2 shows the means and standard deviations of the optimal values of $\lambda$ and $\gamma$.

The following observations are made from Fig. 2: 1) The value of the external-coupling parameter $\lambda$ follows the same trend as the similarity between the multi-task data streams. A large value of $\lambda$ is suitable if the similarity between the multi-task data streams is large. As the similarity between the multi-task data streams gradually decreases and the deviation increases, a smaller value of $\lambda$ becomes more suitable. 2) The value of the internal-coupling parameter $\gamma$ appears to be insensitive to the deviation between the multi-task data streams. $\gamma$ has been investigated in depth in the literature [19,20]. Here, we focus on the external-coupling parameter $\lambda$.

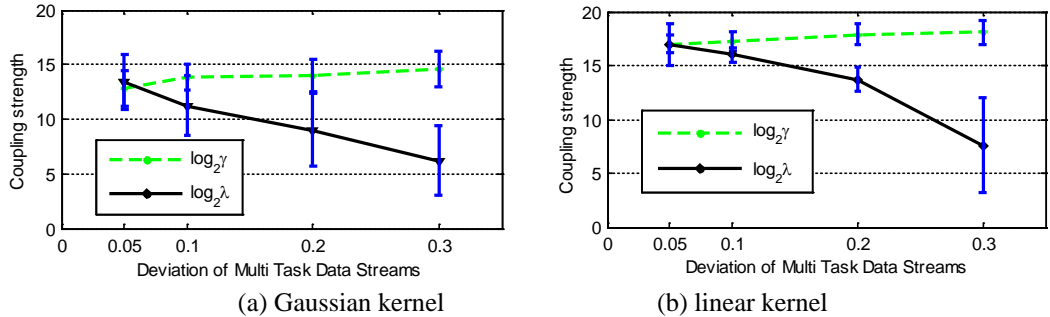

(a) Gaussian kernel  (b) linear kernel
Fig. 2 Optimal values of the coupling parameters $\lambda$ and $\gamma$.

*C. Synthetic Data Streams*

Experiments were conducted to study the performance of DC-SVM on several artificial data streams: DS1, DS2 and the corresponding noisy data streams. For the datasets DS1 and DS2, following the data generation strategy in [19], we independently generated 10 training sets, 10 test datasets and 10 validation datasets for determining the optimal value of $\lambda$. The experimental procedure was similar to that in [19] for TA-SVM, where Gaussian and linear kernels were adopted.

The performances of each method on DS1-Task1 and DS1-Task2 were evaluated under various deviations between the multi-task data streams. Here, $r \in \{0.05, 0.1, 0.2, 0.3\}$ is the deviation of the two data streams. Fig. 3 shows the means and standard deviations of the average test errors on DS1-Task1 and DS1-Task2 over 10 runs of the experiments with Gaussian kernel and Table 2 lists the means and standard deviations of the classification accuracy on DS1 with linear kernel.

The following observations are made from Table 2 and Fig. 3: 1) DC-SVM outperforms TA-SVM and ITA-SVM. 2) As $r$ increases, the relationship between the data streams vanishes. The advantage of DC-SVM decreases; however, it still outperforms TA-SVM and ITA-SVM. 3) The performance of ITA-SVM(Merge) relies heavily on the relationship between the multi-task data streams. When the data streams are highly similar ($r = 0.05$, $r = 0.1$), ITA-SVM(Merge) outperforms TA-SVM and ITA-SVM, while being slightly outperformed by DC-SVM. As the relationship between the data streams weakens, the performance of ITA-SVM(Merge) degrades substantially.

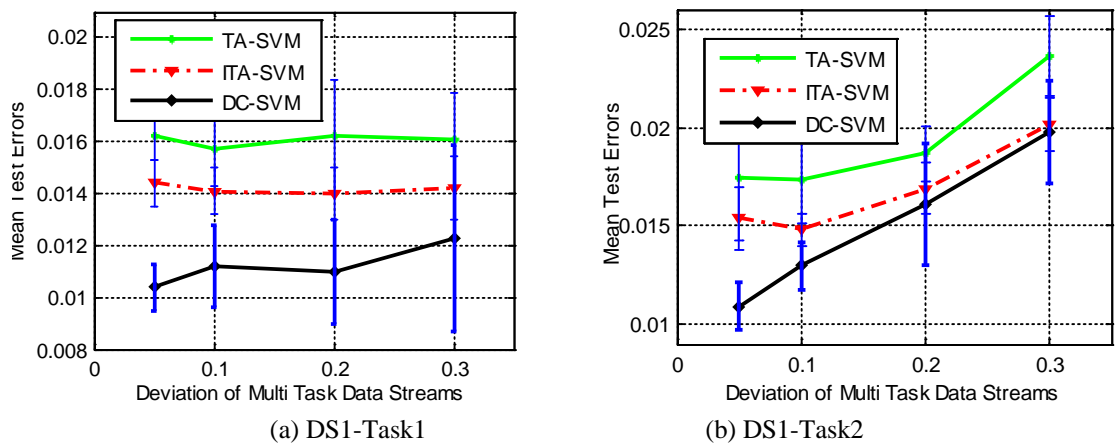

(a) DS1-Task1  (b) DS1-Task2

Fig. 3 Test errors on DS1 with the Gaussian kernel

Table 2

Classification Accuracies (%) on DS1 with linear kernel

| Method | Classification accuracies (%) with different r | | | |
|---|---|---|---|---|
| | r = 0.05 | r = 0.1 | r = 0.2 | r = 0.3 |
| TA-SVM | 97.98±0.34 | 97.94±0.22 | 97.82±0.25 | 97.61±0.32 |
| ITA-SVM | 98.12±0.14 | 98.09±0.14 | 97.99±0.15 | 97.81±0.20 |
| DC-SVM | **98.70±0.28** | **98.63±0.23** | **98.48±0.29** | **98.31±0.32** |
| ITA-SVM(Merge) | 98.45±0.09 | 98.38±0.13 | 98.09±0.13 | 97.80±0.13 |

With the Gaussian kernel are adopted, the average performances of the methods on DS2 are shown in Fig. 4(a). Fig. 4(b) show the average performances of each method on DS3, which has half samples of DS2. Fig. 4 shows that DC-SVM always outperforms the methods that handle the multi-task data streams separately. The advantage decreases as the relationship between the multi-task data streams weakens. According to Fig. 4(a), when the rotation between DS2-Task1 and DS2-Task2 is $2^0$, DC-SVM exhibits a 1% improvement in classification accuracy relative to ITA-SVM. According to Fig. 4(b), when half samples are adopted, DC-SVM exhibits a 2% improvement relative to ITA-SVM with the same $2^0$ rotation between DS3-Task1 and DS3-Task2. Comparing Fig. 4(a) and Fig. 4(b), the proposed multi-task data stream modeling method exhibits a larger performance improvement as the sample size of the data in each time window decreases. The same phenomena can be observed on DS2 and DS3 when linear kernel was adopted, which did not list here for saving space.

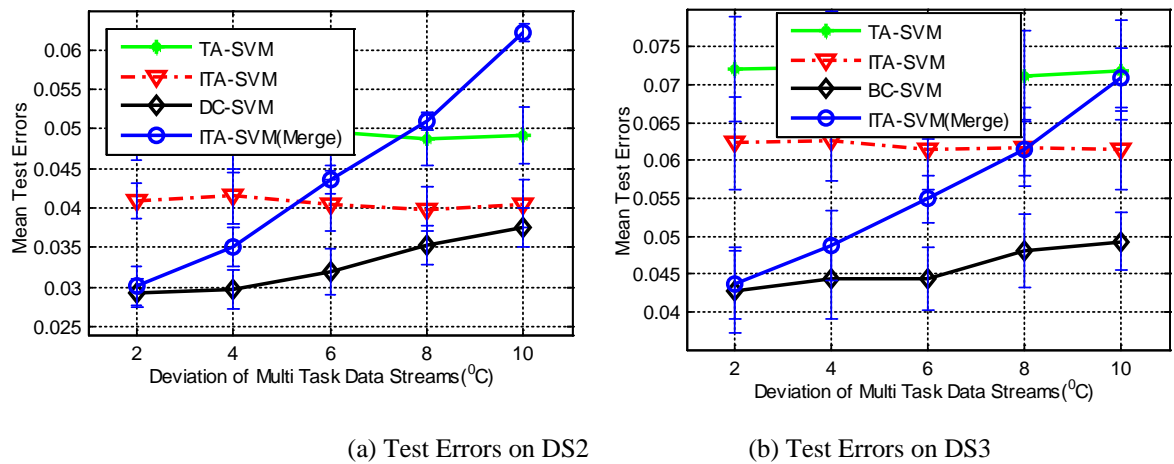

(a) Test Errors on DS2    (b) Test Errors on DS3

Fig. 4 Test errors on DS2 and DS3

In addition, the performances of the methods on noisy datasets DS4 and DS5 were evaluated. With the same settings as in the experiments above, the results are listed in Tables 3 and 4. The results demonstrate that DC-SVM outperforms TA-SVM and ITA-SVM in the noisy environment.

TABLE 3

CLASSIFICATION ACCURACIES (%) ON NOISY DATA STREAM DS4 WITH 10% NOISY DATA.

| Methods | Kernel | Classification accuracies (%) with various diversities | | | |
|---|---|---|---|---|---|
| | | r = 0.05 | r = 0.1 | r = 0.2 | r = 0.3 |
| TA-SVM |  | 88.10±0.25 | 88.07±0.41 | 88.05±0.14 | 87.94±0.38 |
| ITA-SVM | Gauss | 87.77±0.16 | 87.75±0.26 | 87.69±0.25 | 87.55±0.56 |
| DC-SVM |  | **88.58±0.12** | **88.35±0.17** | **88.29±0.19** | **88.14±0.25** |
| TA-SVM |  | 87.46±0.97 | 87.43±0.45 | 87.35±0.59 | 87.31±0.33 |
| ITA-SVM | Linear | 87.23±0.29 | 87.10±0.37 | 86.99±0.30 | 86.88±0.47 |
| DC-SVM |  | **88.17±0.21** | **87.97±0.32** | **87.90±0.20** | **87.73±0.29** |

TABLE 4

CLASSIFICATION ACCURACIES (%) ON NOISY DATA STREAM DS5 WITH 10% NOISY DATA

| Methods | Kernel | Classification accuracies (%) with various diversities | | | |
|---|---|---|---|---|---|
| | | r = 0.05 | r = 0.1 | r = 0.2 | r = 0.3 |
| TA-SVM |  | 83.68±0.73 | 83.73±0.79 | 84.26±0.90 | 84.28±0.58 |
| ITA-SVM | Gauss | 83.95±0.63 | 84.04±0.57 | 84.33±0.70 | 84.19±0.57 |
| DC-SVM |  | **85.67±0.55** | **85.57±0.56** | **85.83±0.41** | **85.42±0.37** |
| TA-SVM |  | 85.57±0.56 | 85.68±0.27 | 85.96±0.42 | 85.61±0.33 |
| ITA-SVM | Linear | 85.30±0.80 | 85.35±0.55 | 85.65±0.48 | 85.51±0.51 |
| DC-SVM |  | **86.72±0.47** | **86.46±0.49** | **86.64±0.50** | **86.22±0.48** |

*D. Gas Sensor Array Drift Dataset*

The proposed method was evaluated on a real-world dataset: a gas sensor array drift dataset [21], noted as GSADD. This dataset contains 13,910 measurements from 16 chemical sensors. The sensors are utilized in simulations of drift compensation in a discrimination task with 6 gases at various levels of concentration. The dataset was collected in 36 months from January 2007 to February 2011 from a gas delivery platform facility. The resulting dataset is comprised of recordings from 6 distinct pure gaseous substances. In this scene the sensor drift may reduce the performance of classifier. This is a typical concept drift problem and the dataset has been regarded as a multi-class dataset in previous research. In this experiment, part of the dataset was selected for binary classification.

Data streams GSADD-Task1 and Task2 were created by selecting ammonia and acetaldehyde,

respectively, as the positive samples. These two types of gases have similar chemical properties. Ammonia has 2565 samples. For acetaldehyde, there were 2926 samples originally, of which 361 samples (no. 2100-2460) were removed to balance the sizes of two data streams. Ethylene has 1936 samples and was selected as the negative sample of GSADD-Task1 and Task2. Data streams GSADD were divided into 25 subsets according to the time sequence, where the data size for each subset was 180 (one additional sample is added to the final subset).

This is a typical prediction task: the former $N$ batches of subsets are used to train a model that is used to predict the $(N+1)th$ subset. The experiment was divided into two steps: First, $N-1$ batches were used to train $N-1$ classifier and the current batch, namely, the $Nth$ subset, was used as the validation set to determine the optimal parameters for each method. Second, $N$ classifiers were trained with the optimized parameters on the training data and the $Nth$ classifier was used to predict the next batch, namely, the $(N+1)th$ batch. The training and prediction processes were repeated until the last batch of data was predicted. The value of $N$ was varied from 3 to 8 to analyze the influence of this parameter. When $N$ exceeded 8, the methods exhibited no substantial performance difference. The average prediction accuracies of each method on the data streams GSADD were evaluated with the same settings as in the above experiments. The performance comparison is plotted in Fig. 5.

The following observations are made from Fig. 5: 1) DC-SVM outperforms TA-SVM and ITA-SVM regardless of the number of batches that are used for training and prediction; 2) When more batches are used for training, the performances of the methods increase gradually, which may be due to the increase in the availability of useful information by utilizing more batches of datasets; 3) The classification performances of DC-SVM and ITA-SVM are optimal when the number of batches that are used for training is 7. Further increasing the number of batches no longer improves the performance.

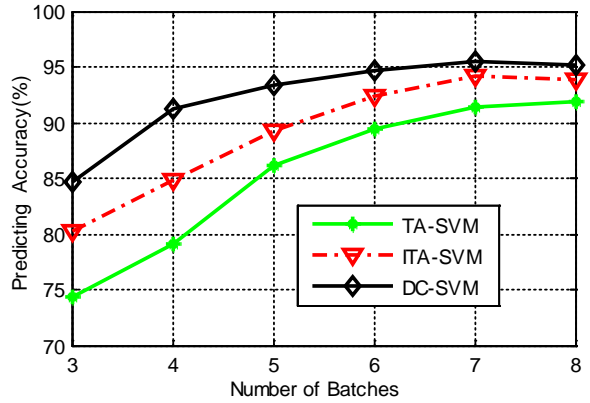

Fig. 5 Prediction accuracy on GSADD

*E. Water Quality Monitoring Data of Taihu Lake*

Surface water automatic monitoring weekly report datasets were collected from the government website of Ministry of Ecological Environment of the P. R. China [47]. Originally, there were three sampling points in Taihu Lake Area and the sampling frequency was once a week, from Jan. 2004 to Dec. 2018. In total, 724 weeks of data were recorded with some weeks unreported. Each dataset has 8 attributes. The first two attributes are the year and the week and the next four are the indicators of pH, DO (mg/l), CODMn (mg/l), and NH3-N (mg/l). The last two attributes are the water quality measurements of this week and the previous week, which are graded from 1 to 6. These are the traditional concept drift datasets. As the sensors may be desensitized and water quality indicators have typical seasonal characteristics, all the properties of the datasets may vary with the seasons. More details of these balanced data streams are given in Table 5.

TABLE 5

DETAILS OF WATER QUALITY DATASETS

| Data stream | Positive samples | Negative samples | Water quality monitoring position |
| --- | --- | --- | --- |
| Water Quality Task1 | 409 | 315 | in Qingpu Rapid Water Port |
| Water Quality Task2 | 349 | 375 | in Lan Shanzui, Yixing |
| Water Quality Task3 | 438 | 286 | in Wang Jiangjing, Jiaxing |

The main objective of our experiments is to predict the water quality according to the values that were collected for four indicators. In our experiments, data streams of water quality, namely, Task1, Task2, and

Task3, were divided into 181 batches with each batch containing 4 samples according to the time sequence.

The first experiment was conducted on Water Quality Task1, Task2, and we use the same experimental settings as in subsection D, where the former $N$ batches of subsets were used to train a model that was used to predict the $(N+1)th$ batch. The value of $N$ is varied from 5 to 25. When $N$ exceeded 20, the performances of the methods did not increase stably or even exhibited a downward trend. The prediction accuracies of each method on the Water Quality Task1 and Task2 were evaluated with the same settings as in the above experiments. The performance is reported in Fig. 6. With the same way, experimental results for Water Quality Task1 and Task3 are plotted in Fig. 7, and experimental results for Water Quality Task2 and Task3 are plotted in Fig. 8. From Figs. 6-8, the similar observations can be obtained as the experimental results in subsection D.

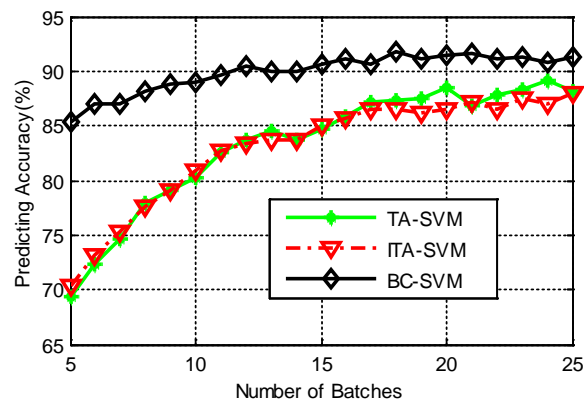

Fig. 6 Prediction accuracy on Water Quality Task1 and Task2.

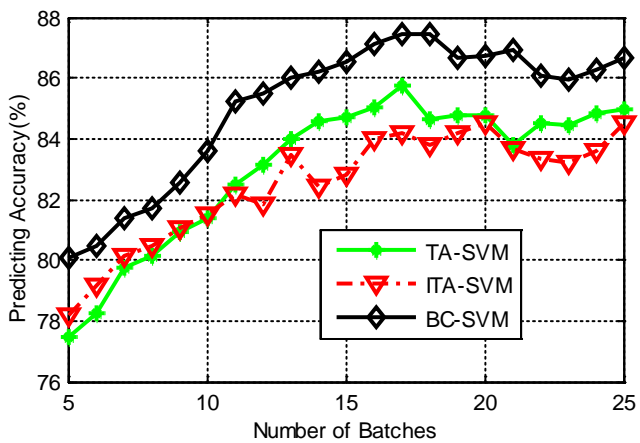 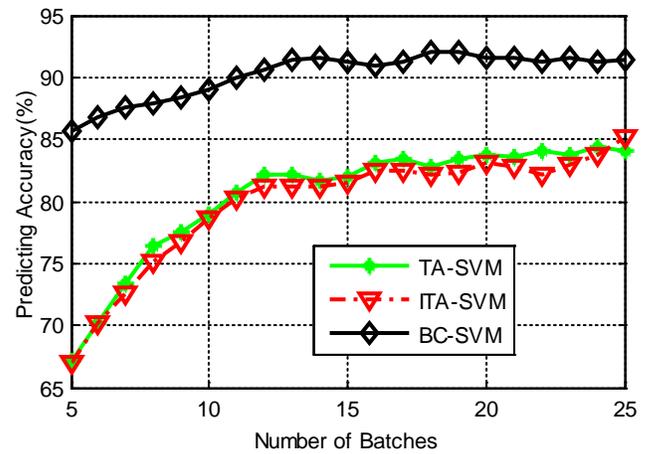

Fig. 7 Prediction accuracies on Water Quality Task1 and Task3.  Fig. 8 Prediction accuracies on Water Quality Task2 and Task3.

## F. Air Quality Dataset

The Air Quality dataset [48] was used by Vito et al to calibrate a chemical sensor [49]. It contains 9358 instances of hourly averaged responses from an array of 5 metal oxide chemical sensors embedded in an air quality chemical multi-sensor device. The device was located in the field in a polluted area, at road level, within an Italian city. Data were recorded from March 2004 to February 2005 (one year) and represent the longest freely available recordings of field-deployed air quality chemical sensor device responses. In the original Air Quality dataset, there are fifteen attributes. The first two are Date and Time. Large amounts of data are missing for four properties (CO(GT), NMHC(GT), NOx(GT), NO2(GT)); hence, only nine attributes were considered in our experiment. For these nine attributes, the missing data are filled via average interpolation according to the values of the previous hour and the next hour. From the chemical perspective, C6H6 and NMHC differ but are of high relevance. Here, we used two properties, namely, C6H6(GT) and S2(NMHC), as the outputs by labeling them as High (above the average) or Low (below the average) and the remaining seven properties were used as inputs. The dataset was divided into 389 batches according to the time sequence, with each batch containing 24 samples that were collected in a specified day, except the last batch, which contains all the remaining samples. When predicting C6H6(GT) and S2(NMHC) levels with TA-SVM or with ITA-SVM, two models are trained separately. For DC-SVM, two models are trained simultaneously for C6H6(GT) and S2(NMHC).

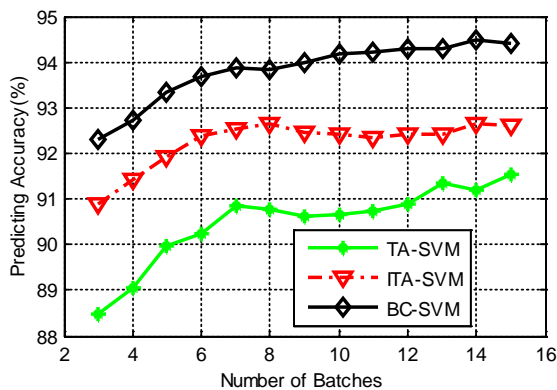

Fig. 10 Prediction accuracies on Air Quality Dataset.

We use the same experimental settings as in subsection D, namely, use the former $N$ batches of subsets to train the model that is used to predict the $(N+1)th$ subset. The value of $N$ is varied from 3 to 15 to evaluate the performance of DC-SVM. The performance comparison is shown in Fig. 9. According to Fig. 9, we can draw the same conclusion as in the above experiments: DC-SVM always performs better than single-task methods for multi-task data streams classification.

## V. CONCLUSIONS

In this paper, a novel method, namely, DC-SVM, is proposed for classifying multi-task data streams by using internal coupling and external coupling simultaneously. Attributed to its ability to leverage the relationships within individual data streams and the relationships among multi-task data streams, DC-SVM outperforms the methods that can only handle each data stream in multi-task data streams independently. The experimental results demonstrate the promising performance of the proposed method. Nevertheless, there remains room for further improvement, for example, in determining 1) how to implement the adaptive modeling method when the samples in the data streams are unbalanced; 2) how to ensure that the data streams can uniformly benefit from each other in a balanced manner; and 3) how to effectively and efficiently train the model for multi-task data streams that contain more than two data streams. Further investigations will be carried out to study these topics in detail.

### ACKNOWLEDGEMENTS

This work was supported in part by the National Key Research Program of China (2016YFB0800803), by the National Natural Science Foundation of China (61772239), the National First-Class Discipline Program of Light Industry Technology and Engineering under Grant LITE2018-02, the Outstanding Youth Fund of Jiangsu Province (BK20140001), the Hong Kong Research Grants Council (PolyU 152040/16E) and the Hong Kong Polytechnic University (G-UA68, G-UA3W), the Open Foundation of Jiangsu Key Laboratory of Media Design and Software Technology (18ST0203).